\documentclass[
]{ceurart}

\usepackage{listings}
\usepackage[acronym]{glossaries}
\makeglossaries
\newacronym{qa}{QA}{Question Answering}
\newacronym{ir}{IR}{Information Retrieval}
\newacronym{llm}{LLM}{Large Language Model}
\newacronym{rag}{RAG}{Retrieval-Augmented Generation}
\newacronym{pse}{PSE}{Programmable Search Engine}
\newacronym{pmbm}{PMBM}{PubMed Best Match}
\newacronym{moe}{MoE}{Mixture of Experts}
\newacronym{nlp}{NLP}{Natural Language Processing}

\usepackage{tikz}
\usetikzlibrary{shapes.geometric, arrows}

\tikzstyle{startstop} = [rectangle, rounded corners, minimum width=3cm, minimum height=1cm,text centered, draw=black]
\tikzstyle{process} = [rectangle, minimum width=3cm, minimum height=1cm, text centered, draw=black]
\tikzstyle{decision} = [diamond, minimum width=3cm, minimum height=1cm,text centered, draw=black]
\tikzstyle{arrow} = [thick,->]

\usepackage{booktabs}

\begin{document}

\copyrightyear{2025}
\copyrightclause{Copyright for this paper by its authors.
  Use permitted under Creative Commons License Attribution 4.0
  International (CC BY 4.0).}

\conference{CLEF 2025 Working Notes, 9 -- 12 September 2025, Madrid, Spain}

\title{LLM Ensemble for RAG: Role of Context Length in Zero-Shot Question Answering for BioASQ Challenge}

\title[mode=sub]{Notebook for the BioASQ Lab at CLEF 2025}


\author[1]{Dima Galat}[
orcid=0000-0003-3825-2142,
email=dima.galat [@] student.uts.edu.au,
]
\address[1]{University of Technology Sydney (UTS), Australia}

\author[2]{Diego Molla-Aliod}[
orcid=0000-0003-4973-0963,
email=diego.molla-aliod [@] mq.edu.au,
]
\address[2]{Macquarie University, Australia}

\begin{abstract}
Biomedical question answering (QA) poses significant challenges due to the need for precise interpretation of specialized knowledge drawn from a vast, complex, and rapidly evolving corpus. In this work, we explore how large language models (LLMs) can be used for information retrieval (IR), and an ensemble of zero-shot models can accomplish state-of-the-art performance on a domain-specific Yes/No QA task. Evaluating our approach on the BioASQ challenge tasks, we show that ensembles can outperform individual LLMs and in some cases rival or surpass domain-tuned systems - all while preserving generalizability and avoiding the need for costly fine-tuning or labeled data. Our method aggregates outputs from multiple LLM variants, including models from Anthropic and Google, to synthesize more accurate and robust answers. Moreover, our investigation highlights a relationship between context length and performance: while expanded contexts are meant to provide valuable evidence, they simultaneously risk information dilution and model disorientation. These findings emphasize IR as a critical foundation in Retrieval-Augmented Generation (RAG) approaches for biomedical QA systems. Precise, focused retrieval remains essential for ensuring LLMs operate within relevant information boundaries when generating answers from retrieved documents. Our results establish that ensemble-based zero-shot approaches, when paired with effective RAG pipelines, constitute a practical and scalable alternative to domain-tuned systems for biomedical question answering.
\end{abstract}

\begin{keywords}
  large language model \sep
  biomedical question answering \sep
  information retrieval \sep
  natural language processing \sep
  BioASQ \sep
  CEUR-WS
\end{keywords}

\maketitle

\section{Introduction}

Biomedical \gls{qa} is a challenging task, and BioASQ is an annual competition aimed at fostering the development of intelligent systems specialized in \gls{ir} and \gls{qa} within the biomedical domain \citep{Tsatsaronis2015}. The competition consists of three distinct phases: phase A, which focuses on a biomedical \gls{ir} task; phase B, centered on \gls{qa} and summarization tasks; and phase A+, which seeks to develop an end-to-end approach for both phases of the challenge combined, has been introduced for the first time in 2024 to encourage further development in this area \citep{bioasq12}. Notably, a third of the participating teams attempted all three phases of the challenge that year.\newline\\
\textbf{This paper presents the following key advancements:}

\begin{itemize}

\item We develop a zero-shot \gls{qa} ensembling framework that leverages \glspl{llm} and answer synthesis to accomplish state-of-the-art results on a Yes/No \gls{qa} task.

\item We design a multi-stage biomedical \gls{ir} pipeline that combines \gls{llm}-generated queries, BM25 lexical search, and semantic reranking to enable high-recall, high-precision document selection for \gls{rag} \gls{qa} in Phase A+.

\item We demonstrate how context influences the \gls{qa} results and show that \gls{ir} is still an important part of a modern end-to-end \gls{qa} solution.

\end{itemize}
\section{Related work}

\gls{ir} and \gls{qa} are two well-established fields in \gls{nlp} and both of them have had extensive research for decades. Most relevant for us is work on the use of \glspl{llm} for query generation in \gls{ir}, answer re-ranking after a first pass by a traditional retrieval system, and the use of \glspl{llm} and \gls{rag} for \gls{qa}.

\glspl{llm} have been used to generate search queries, or to rewrite initial search queries to improve their performance. Techniques include the use of zero-shot, few-shot and chain-of-thought prompting \cite{Jagerman2023,Wang2023,Alaofi2023}, supervised fine-tuning \cite{Peng2024}, and reinforcement learning \cite{Ma2023}.

Search results have been reranked using a wide range of techniques, the most common being applying a similarity metric such as cosine similarity to the output of a cross-encoder or a twin model such as Sentence-BERT \cite{Reimers2019}. Google released the ReFr open-source framework for re-ranking search results, which allows researchers to explore multiple features and learning methods \cite{Google2013}.

\glspl{llm} have also been used to generate answers to questions, either in a zero-shot, or few-shot manner, or by fine-tuning on question-answering datasets. The integration of contextual information through \gls{rag} consistently improves answer quality and factual accuracy compared to \gls{llm} approaches not relying on additional relevant context information \citep{Lewis2020}. A quick look at the proceedings of BioASQ 12 at CLEF 2024 shows that, out of a total of 23 papers, 6 papers used the terms \gls{llm} or \glsentrylong{llm} in their title, and 3 additional papers used the terms \gls{rag} or \glsentrylong{rag}. 
\section{Our methodology}

\subsection{Information Retrieval Pipeline}
\label{chap:ir}
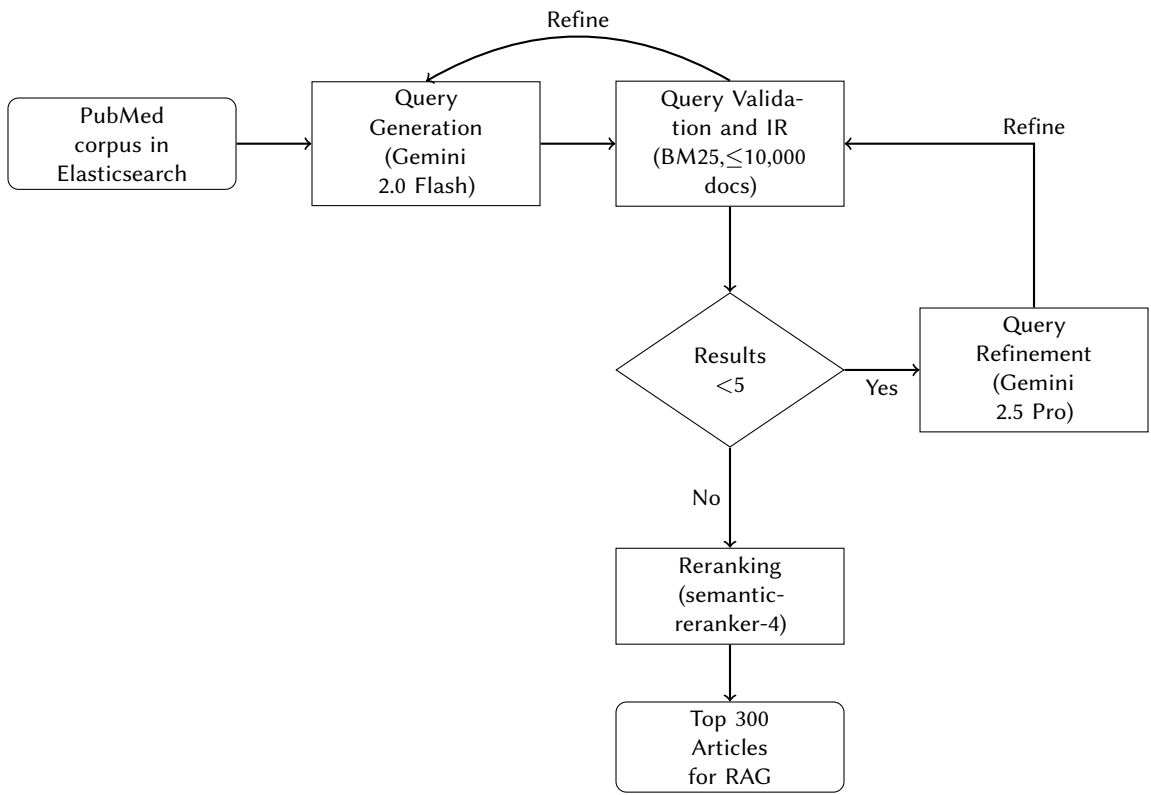
\begin{figure}[h]
  \footnotesize
  \centering
\begin{tikzpicture}[node distance=2cm]

  \node (start) [startstop,text width=2cm] {PubMed corpus in Elasticsearch};
  \node (querygen) [process,right of=start,xshift=2cm,text width=2cm] {Query Generation  (Gemini 2.0 Flash)};
  \node (queryval) [process,right of=querygen,xshift=2cm,text width=2cm] {Query Validation and IR (BM25,$ \leq $10,000 docs)};
  \node (decision) [decision,right of=querygen,xshift=2cm,yshift=-3cm,text width=1cm] {Results $<$5};
  \node (refine) [process,right of=decision,xshift=2cm,text width=2cm] {Query Refinement  (Gemini 2.5 Pro)};
  \node (rerank) [process,right of=querygen,yshift=-6 cm,xshift=2cm,text width=2cm]{Reranking (semantic-reranker-4)};
  \node (end) [startstop,right of=querygen,yshift=-8 cm,xshift=2cm,text width=2cm]{Top 300 Articles for RAG};
  
  \draw[arrow] (start)--(querygen);
  \draw[arrow] (querygen)--(queryval);
  \draw[arrow] (queryval) -- (decision);
  \draw[arrow] (decision) -- node[anchor=east]{No}(rerank);
  \draw[arrow] (rerank) -- (end);
  \draw[arrow] (decision) -- node[anchor=north]{Yes} (refine);
  \draw[arrow] (refine) |- node[anchor=south]{Refine} (queryval);
  \draw[arrow, bend right=30] (queryval.north) to node[above] {Refine} (querygen.north);
  
  \end{tikzpicture}
  \caption{IR process}
  \label{fig:ir}
\end{figure}

To support high-quality \gls{rag} for Phase A+, we developed an \gls{ir} pipeline that integrates traditional lexical search with \gls{llm}-based query generation and semantic reranking (Fig.~\ref{fig:ir}).

We index all PubMed article titles and abstracts in an Elasticsearch instance, using BM25 retrieval as the ranking function. For each input question, we use Gemini 2.0 Flash to generate a structured Elasticsearch query that captures the semantic intent of the question using synonyms, related terms, and full boolean query string syntax rules supported by Elasticsearch. This query is validated using regular expressions and then is used to retrieve up to 10,000 documents.

If the initial query returns fewer than five documents, we invoke Gemini 2.5 Pro Preview (05-06) to automatically revise the query. The model is prompted to enhance retrieval recall by enabling approximate matching and omitting overly rare or domain-specific terms. This refinement step is done to improve the query coverage while maintaining relevance. Our experiments have shown that this process is required in less than 5\% of the queries in the BioASQ 13 test set.

Following document retrieval, we apply a semantic reranking model (Google semantic-ranker-default-004) to reduce the number of candidate documents \citep{GoogleCloud2024}.  This model re-scores the initially retrieved documents based on semantic similarity to the original question, allowing us to select the top 300 most relevant documents. This reranked subset is used for downstream \gls{rag}-based  \gls{qa}, since despite really long context supported by modern Transformer architectures \cite{Zaheer2020, Beltagy2020}, we could not get adequate \gls{qa} results on full article abstracts without this step.

This multi-stage retrieval approach, combining \gls{llm}-generated queries, a traditional BM25 search, and semantic reranking, enables flexible, high-recall, and high-precision document selection tailored to complex biomedical queries.

Finally, we have added additional \gls{ir} searches to handle the cases where a \gls{qa} step does not return a response based on the evidence retrieved from Elasticsearch. We have observed that Elasticsearch context might not provide sufficient evidence for \gls{qa} in 3-7\% of test cases for Phase A+, depending on the batch. An automated process is used to expand \gls{ir} sources to address these cases. First, we are using a Google search restricted to PubMed sources to attempt to find new matches. If that fails, we extend our sources to include Home of the Office of Health Promotion and Disease Prevention, WebMD, Healthline, and Wikipedia. This ensures that we have an answer candidate for all questions in Phase A+ test sets. 

\subsection{Question Answering Pipeline}

We adopt a unified, zero-shot \gls{qa} framework for both Phase A+ and Phase B of the challenge. While the core \gls{qa} procedure remains consistent across phases, Phase A+ incorporates an additional \gls{ir} step to verify the presence of candidate answers within relevant documents (described at the end of Section~\ref{chap:ir}). This ensures that selected documents contain sufficient information to support answer generation.

The system uses zero-shot prompting, tailored to the question type: Yes/No, Factoid, or List. We experiment with multiple types of input context: (1) IR-derived results from Phase A+, (2) curated snippets provided in Phase B, and (3) full abstracts of articles selected during Phase B. This allows us to examine the influence of context granularity on answer accuracy and completeness.

To generate candidate answers, we leverage several large language models (LLMs): Gemini 2.0 Flash, Gemini 2.5 Flash Preview (2025-04-17), and Claude 3.7 Sonnet (2025-02-19). Prompts are adjusted using examples derived from the BioASQ 11 test set, improving the response structure and quality.

To consolidate candidate answers, we perform a secondary synthesis step using Gemini 2.0 Flash. This model is prompted to resolve any contradictions, select the most precise and specific answer components, and integrate complementary information into a single, unified response. As part of this step, the model also returns a confidence score estimating the reliability of the synthesized answer. If the score is below a predefined threshold (0.5, determined empirically), the synthesis is re-run with reduced sampling temperature (from 0.1 to 0.0) to improve determinism. This synthesis process is evaluated using the BioASQ 12 dataset to ensure consistency with benchmark standards.

\section{Results}
\label{res}

\subsection{Phase A+}
BioASQ competition results are evaluated using accuracy, which measures the proportion of correctly answered yes/no questions, and MRR (Mean Reciprocal Rank), which assesses how early correct answers appear in the ranked list of factoid question responses.

Our \gls{ir} pipeline has been actively developed during the BioASQ 13 competition, and as a result, it was not ready in time to submit the first two batches. Notably, on batch 4, our system achieved state-of-the-art results on Yes/No questions, underscoring the effectiveness of the \gls{rag} approach described in Section~\ref{chap:ir} (Table~\ref{tab:phase_a_plus_yes_no}). Despite this success, our system encountered challenges in producing the structured output required for List and Factoid questions, which is a consistent issue we've seen with zero-shot generation. Table~\ref{tab:phase_a_factoid} shows the results for factoid questions. It's important note that Simple truncation is a system using Gemini 2.0 Flash described in Figure~\ref{fig:ir}, and Extractive is the system using Claude 3.7 Sonnet \textemdash which worked better for long context extraction despite technically having a smaller context size available.

\begin{table}
    \caption{Results of our runs on BioASQ 13 Phase A+, Yes/No questions.}
    \label{tab:phase_a_plus_yes_no}
\centering
\begin{tabular}{llll}
\toprule
\textbf{Batch} & \textbf{System} & \textbf{Accuracy} & \textbf{Ranking} \\
\midrule
3 & Extractive & 0.73 & 41\\
 & (last) & 0.23 & 58\\
\midrule
4 & Extractive & 0.92 & 1\\
 & Simple truncation & 0.88 & 11\\
 & Kmeans & 0.65 & 67\\
 & (last) & 0.65 & 67\\
\bottomrule
\end{tabular}
\end{table}

\begin{table}[]
    \caption{Results of our runs on BioASQ 13 Phase A+, Factoid questions.}
\label{tab:phase_a_factoid}
\centering
\begin{tabular}{llll}
\toprule
\textbf{Batch} & \textbf{System} & \textbf{MRR} & \textbf{Ranking} \\
\midrule
3 & Extractive & 0.14 & 41\\
 & (last) & 0.05 & 47\\
\midrule
4 & Extractive & 0.43 & 17\\
 & Simple truncation & 0.29 & 51\\
 & Kmeans & 0.05 & 62\\
 & (last) & 0.05 & 62\\
\bottomrule
\end{tabular}
\end{table}

\subsection{Phase  B}
We evaluated our system on the BioASQ 12 dataset and observed competitive performance across all batches. Overall, our system ranked 5th for exact answers across our experiments compared to systems presented last year, with batch 4 posing the greatest challenge across all question types. Among the models tested, Gemini consistently outperformed Claude, with Gemini 2.0 Flash showing significantly better results compared to Gemini 2.5 Flash or Pro. The system showed its strongest performance on Yes/No questions, while List-type questions were the most challenging. We found that using longer contexts \textemdash like full abstracts \textemdash generally hurts the answer quality. This can be attributed to difficulties in generating well-structured outputs for List and Factoid questions, further highlighting that answer generation and formatting should be decoupled into separate stages to improve the results on exact questions requiring more nuanced responses than a binary answer. 

Across the BioASQ 13 Phase B batches, our system demonstrated consistently strong performance on Yes/No questions. In batches 1, 2, and 3 we achieved state-of-the-art results on Yes/No questions using snippets provided for this stage and an ensemble of Gemini 2.0 Flash and Gemini 2.5 Flash Preview (2025-04-17) (Table~\ref{tab:phase_b_yes_no}). In batch 3, we managed to refine our prompts to also achieve a second place in Factoid questions using the same model and context selection (Table~\ref{tab:phase_b_factoid}). In batch 4, this approach ranked third on Yes/No questions (Table~\ref{tab:phase_b_yes_no}), demonstrating the robustness of our synthesis pipeline across varied input selection strategies.  

\begin{table}
    \caption{Results of our runs on BioASQ 13 Phase B, Yes/No questions.}
    \label{tab:phase_b_yes_no}
\centering
\begin{tabular}{llll}
\toprule
\textbf{Batch} & \textbf{System} & \textbf{Accuracy} & \textbf{Ranking} \\
\midrule
1 & Simple truncation & 1.00 & 1\\
 & Extractive & 0.94 & 23\\
 & Kmeans & 0.94 & 23\\
 & (last) & 0.29 & 72\\
\midrule
2 & Simple truncation & 1.00 & 1\\
 & Extractive & 0.88 & 47\\
 & Kmeans & 0.88 & 47\\
 & (last) & 0.47 & 72\\
\midrule
3 & Simple truncation & 0.95 & 1\\
 & Extractive & 0.95 & 1\\
 & Kmeans & 0.95 & 1\\
 & (last) & 0.27 & 66\\
\midrule
4 & Simple truncation & 0.96 & 3\\
 & Extractive & 0.96 & 3\\
 & Kmeans & 0.96 & 3\\
 & (last) & 0.27 & 79\\
\bottomrule
\end{tabular}
\end{table}

\begin{table}
    \caption{Results of our runs on BioASQ 13 Phase B, Factoid questions.}
    \label{tab:phase_b_factoid}
\centering
\begin{tabular}{llll}
\toprule
\textbf{Batch} & \textbf{System} & \textbf{MRR} & \textbf{Ranking} \\
\midrule
1 & Simple truncation & 0.44 & 38\\
 & Extractive & 0.38 & 53\\
 & Kmeans & 0.37 & 54\\
 & (last) & 0.11 & 67\\
\midrule
2 & Simple truncation & 0.57 & 20\\
 & KMeans & 0.54 & 32\\
 & Extractive & 0.52 & 36\\
 & (last) & 0.07 & 66\\
\midrule
3 & Simple truncation & 0.60 & 2\\
 & Kmeans & 0.49 & 4\\
 & Extractive & 0.10 & 55\\
 & (last) & 0.03 & 59\\
\midrule
4 & Kmeans & 0.50 & 30\\
 & Extractive & 0.48 & 38\\
 & Simple truncation & 0.45 & 46\\
 & (last) & 0.05 & 73\\
\bottomrule
\end{tabular}
\end{table}

\section{Conclusion and Future Work}

\subsection{Conclusions}

This work demonstrates the effectiveness of integrating zero-shot \glspl{llm} with traditional \gls{ir} systems for providing an end-to-end approach for \gls{qa}. \glspl{llm} can help bridge the semantic gap between queries and relevant documents and are effective when used in a specialized domain \citep{Karpukhin2020}. Our findings align with the growing trend, as evidenced by the substantial adoption of \gls{llm} and \gls{rag} techniques in recent BioASQ competitions \citep{bioasq12}, as well as with studies that show that query rewriting provides substantial performance advantages \cite{Jagerman2023,Alaofi2023,Wang2023}. 

 The multi-stage \gls{ir} pipeline approach provides robust performance while maintaining computational efficiency. We have shown that a combination of \gls{llm} query generation, cross-encoder re-ranking, and \gls{rag} are capable of processing very long domain-specific contexts, achieving a SOTA Yes/No \gls{qa} system performance on multiple BioASQ batches this year. We found that producing structured data outputs for other question types can be challenging, especially as the context size increases. 

\subsection{Future Work}

Several promising directions emerge from this research that warrant further investigation:

\textbf{Evaluation and Robustness}: Developing more comprehensive evaluation frameworks that assess not only accuracy but also consistency, bias, and hallucination rates across diverse query types and domains \cite{factscore}. ARES proposes an approach for evaluating RAG systems that fine-tunes lightweight \gls{llm} judges using synthetic data and minimal human annotations, achieving high accuracy across tasks and domains while outperforming prior methods \cite{Falcon2024}.

\textbf{Interactive Systems}: Development of conversational interfaces that can clarify ambiguous queries, and provide explanations for retrieved information and generated answers \cite{Qu2020}. Uncertainty detection methods can be used to dynamically trigger retrieval in \gls{rag} systems, reducing unnecessary retrievals while maintaining or even improving answer quality in long-form question answering tasks \citep{zhang2025uncertainty}.

\bibliography{sample-1col}

\begin{thebibliography}{18}
\expandafter\ifx\csname natexlab\endcsname\relax\def\natexlab#1{#1}\fi
\providecommand{\url}[1]{\texttt{#1}}
\providecommand{\href}[2]{#2}
\providecommand{\path}[1]{#1}
\providecommand{\DOIprefix}{doi:}
\providecommand{\ArXivprefix}{arXiv:}
\providecommand{\URLprefix}{URL: }
\providecommand{\Pubmedprefix}{pmid:}
\providecommand{\doi}[1]{\href{http://dx.doi.org/#1}{\path{#1}}}
\providecommand{\Pubmed}[1]{\href{pmid:#1}{\path{#1}}}
\providecommand{\bibinfo}[2]{#2}
\ifx\xfnm\relax \def\xfnm[#1]{\unskip,\space#1}\fi
\bibitem[{Tsatsaronis et~al.(2015)Tsatsaronis, Balikas, Malakasiotis, Partalas, Zschunke, Alvers, Weissenborn, Krithara, Petridis, Polychronopoulos, Almirantis, Pavlopoulos, Baskiotis, Gallinari, Artiéres, Ngomo, Heino, Gaussier, Barrio-Alvers, Schroeder, Androutsopoulos, and Paliouras}]{Tsatsaronis2015}
\bibinfo{author}{G.~Tsatsaronis}, \bibinfo{author}{G.~Balikas}, \bibinfo{author}{P.~Malakasiotis}, \bibinfo{author}{I.~Partalas}, \bibinfo{author}{M.~Zschunke}, \bibinfo{author}{M.~R. Alvers}, \bibinfo{author}{D.~Weissenborn}, \bibinfo{author}{A.~Krithara}, \bibinfo{author}{S.~Petridis}, \bibinfo{author}{D.~Polychronopoulos}, \bibinfo{author}{Y.~Almirantis}, \bibinfo{author}{J.~Pavlopoulos}, \bibinfo{author}{N.~Baskiotis}, \bibinfo{author}{P.~Gallinari}, \bibinfo{author}{T.~Artiéres}, \bibinfo{author}{A.~C.~N. Ngomo}, \bibinfo{author}{N.~Heino}, \bibinfo{author}{E.~Gaussier}, \bibinfo{author}{L.~Barrio-Alvers}, \bibinfo{author}{M.~Schroeder}, \bibinfo{author}{I.~Androutsopoulos}, \bibinfo{author}{G.~Paliouras},
\newblock \bibinfo{title}{An overview of the {BioASQ} large-scale biomedical semantic indexing and question answering competition},
\newblock \bibinfo{journal}{BMC Bioinformatics} \bibinfo{volume}{16} (\bibinfo{year}{2015}). \DOIprefix\doi{10.1186/s12859-015-0564-6}.
\bibitem[{Nentidis et~al.(2024)Nentidis, Krithara, Paliouras, Krallinger, Sanchez, Lima, Farre, Loukachevitch, Davydova, and Tutubalina}]{bioasq12}
\bibinfo{author}{A.~Nentidis}, \bibinfo{author}{A.~Krithara}, \bibinfo{author}{G.~Paliouras}, \bibinfo{author}{M.~Krallinger}, \bibinfo{author}{L.~G. Sanchez}, \bibinfo{author}{S.~Lima}, \bibinfo{author}{E.~Farre}, \bibinfo{author}{N.~Loukachevitch}, \bibinfo{author}{V.~Davydova}, \bibinfo{author}{E.~Tutubalina}, \bibinfo{title}{BioASQ at CLEF2024: The Twelfth Edition of the Large-Scale Biomedical Semantic Indexing and Question Answering Challenge}, \bibinfo{year}{2024}, pp. \bibinfo{pages}{490--497}. \DOIprefix\doi{10.1007/978-3-031-56069-9_67}.
\bibitem[{Jagerman et~al.(2023)Jagerman, Zhuang, Qin, Wang, and Bendersky}]{Jagerman2023}
\bibinfo{author}{R.~Jagerman}, \bibinfo{author}{H.~Zhuang}, \bibinfo{author}{Z.~Qin}, \bibinfo{author}{X.~Wang}, \bibinfo{author}{M.~Bendersky},
\newblock \bibinfo{title}{Query expansion by prompting large language models},
\newblock \bibinfo{journal}{arXiv preprint arXiv:2305.03653}  (\bibinfo{year}{2023}).
\bibitem[{Wang et~al.(2023)Wang, Yang, and Wei}]{Wang2023}
\bibinfo{author}{L.~Wang}, \bibinfo{author}{N.~Yang}, \bibinfo{author}{F.~Wei},
\newblock \bibinfo{title}{Query2doc: Query expansion with large language models},
\newblock \bibinfo{journal}{arXiv preprint arXiv:2303.07678}  (\bibinfo{year}{2023}).
\bibitem[{Alaofi et~al.(2023)Alaofi, Gallagher, Sanderson, Scholer, and Thomas}]{Alaofi2023}
\bibinfo{author}{M.~Alaofi}, \bibinfo{author}{L.~Gallagher}, \bibinfo{author}{M.~Sanderson}, \bibinfo{author}{F.~Scholer}, \bibinfo{author}{P.~Thomas},
\newblock \bibinfo{title}{Can generative {LLMs} create query variants for test collections? an exploratory study},
\newblock in: \bibinfo{booktitle}{Proceedings of the 46th international ACM SIGIR conference on research and development in information retrieval}, \bibinfo{year}{2023}, pp. \bibinfo{pages}{1869--1873}.
\bibitem[{Peng et~al.(2024)Peng, Li, Jiang, Wang, Ou, Zeng, Xu, Xu, and Chen}]{Peng2024}
\bibinfo{author}{W.~Peng}, \bibinfo{author}{G.~Li}, \bibinfo{author}{Y.~Jiang}, \bibinfo{author}{Z.~Wang}, \bibinfo{author}{D.~Ou}, \bibinfo{author}{X.~Zeng}, \bibinfo{author}{D.~Xu}, \bibinfo{author}{T.~Xu}, \bibinfo{author}{E.~Chen},
\newblock \bibinfo{title}{Large language model based long-tail query rewriting in taobao search},
\newblock in: \bibinfo{booktitle}{Companion Proceedings of the ACM Web Conference 2024}, \bibinfo{year}{2024}, pp. \bibinfo{pages}{20--28}.
\bibitem[{Ma et~al.(2023)Ma, Gong, He, Zhao, and Duan}]{Ma2023}
\bibinfo{author}{X.~Ma}, \bibinfo{author}{Y.~Gong}, \bibinfo{author}{P.~He}, \bibinfo{author}{H.~Zhao}, \bibinfo{author}{N.~Duan},
\newblock \bibinfo{title}{Query rewriting in retrieval-augmented large language models},
\newblock in: \bibinfo{booktitle}{Proceedings of the 2023 Conference on Empirical Methods in Natural Language Processing}, \bibinfo{year}{2023}, pp. \bibinfo{pages}{5303--5315}.
\bibitem[{Reimers and Gurevych(2019)}]{Reimers2019}
\bibinfo{author}{N.~Reimers}, \bibinfo{author}{I.~Gurevych},
\newblock \bibinfo{title}{Sentence-{BERT}: Sentence embeddings using {S}iamese {BERT}-networks},
\newblock in: \bibinfo{editor}{K.~Inui}, \bibinfo{editor}{J.~Jiang}, \bibinfo{editor}{V.~Ng}, \bibinfo{editor}{X.~Wan} (Eds.), \bibinfo{booktitle}{Proceedings of the 2019 Conference on Empirical Methods in Natural Language Processing and the 9th International Joint Conference on Natural Language Processing (EMNLP-IJCNLP)}, \bibinfo{publisher}{Association for Computational Linguistics}, \bibinfo{address}{Hong Kong, China}, \bibinfo{year}{2019}, pp. \bibinfo{pages}{3982--3992}. \URLprefix \url{https://aclanthology.org/D19-1410/}. \DOIprefix\doi{10.18653/v1/D19-1410}.
\bibitem[{Bikel and Hall(2013)}]{Google2013}
\bibinfo{author}{D.~M. Bikel}, \bibinfo{author}{K.~B. Hall},
\newblock \bibinfo{title}{{ReFr}: An open-source reranker framework},
\newblock in: \bibinfo{booktitle}{Interspeech 2013}, \bibinfo{year}{2013}, pp. \bibinfo{pages}{756--758}.
\bibitem[{Lewis et~al.(2020)Lewis, Perez, Piktus, Petroni, Karpukhin, Goyal, Küttler, Lewis, Yih, Rocktäschel et~al.}]{Lewis2020}
\bibinfo{author}{P.~Lewis}, \bibinfo{author}{E.~Perez}, \bibinfo{author}{A.~Piktus}, \bibinfo{author}{F.~Petroni}, \bibinfo{author}{V.~Karpukhin}, \bibinfo{author}{N.~Goyal}, \bibinfo{author}{H.~Küttler}, \bibinfo{author}{M.~Lewis}, \bibinfo{author}{W.-t. Yih}, \bibinfo{author}{T.~Rocktäschel}, et~al.,
\newblock \bibinfo{title}{Retrieval-augmented generation for knowledge-intensive {NLP} tasks},
\newblock in: \bibinfo{booktitle}{Advances in Neural Information Processing Systems}, volume~\bibinfo{volume}{33}, \bibinfo{year}{2020}, pp. \bibinfo{pages}{9459--9474}.
\bibitem[{{Google Cloud}(2024)}]{GoogleCloud2024}
\bibinfo{author}{{Google Cloud}}, \bibinfo{title}{Ranking and re-ranking search results}, \bibinfo{howpublished}{\url{https://cloud.google.com/generative-ai-app-builder/docs/ranking}}, \bibinfo{year}{2024}. \bibinfo{note}{Accessed: 2025}.
\bibitem[{Zaheer et~al.(2020)Zaheer, Guruganesh, Dubey, Ainslie, Alberti, Ontanon, Pham, Ravula, Wang, Yang, and Ahmed}]{Zaheer2020}
\bibinfo{author}{M.~Zaheer}, \bibinfo{author}{G.~Guruganesh}, \bibinfo{author}{A.~Dubey}, \bibinfo{author}{J.~Ainslie}, \bibinfo{author}{C.~Alberti}, \bibinfo{author}{S.~Ontanon}, \bibinfo{author}{P.~Pham}, \bibinfo{author}{A.~Ravula}, \bibinfo{author}{Q.~Wang}, \bibinfo{author}{L.~Yang}, \bibinfo{author}{A.~Ahmed},
\newblock \bibinfo{title}{Big bird: Transformers for longer sequences},
\newblock in: \bibinfo{booktitle}{Advances in Neural Information Processing Systems}, volume \bibinfo{volume}{2020-December}, \bibinfo{year}{2020}.
\bibitem[{Beltagy et~al.(2020)Beltagy, Peters, and Cohan}]{Beltagy2020}
\bibinfo{author}{I.~Beltagy}, \bibinfo{author}{M.~E. Peters}, \bibinfo{author}{A.~Cohan},
\newblock \bibinfo{title}{Longformer: The long-document transformer}  (\bibinfo{year}{2020}).
\bibitem[{Karpukhin et~al.(2020)Karpukhin, Oguz, Min, Lewis, Wu, Edunov, Chen, and Yih}]{Karpukhin2020}
\bibinfo{author}{V.~Karpukhin}, \bibinfo{author}{B.~Oguz}, \bibinfo{author}{S.~Min}, \bibinfo{author}{P.~Lewis}, \bibinfo{author}{L.~Wu}, \bibinfo{author}{S.~Edunov}, \bibinfo{author}{D.~Chen}, \bibinfo{author}{W.-t. Yih},
\newblock \bibinfo{title}{Dense passage retrieval for open-domain question answering},
\newblock in: \bibinfo{booktitle}{Proceedings of the 2020 Conference on Empirical Methods in Natural Language Processing}, \bibinfo{year}{2020}, pp. \bibinfo{pages}{6769--6781}.
\bibitem[{Min et~al.(2023)Min, Krishna, Lyu, Lewis, Yih, Koh, Iyyer, Zettlemoyer, and Hajishirzi}]{factscore}
\bibinfo{author}{S.~Min}, \bibinfo{author}{K.~Krishna}, \bibinfo{author}{X.~Lyu}, \bibinfo{author}{M.~Lewis}, \bibinfo{author}{W.-t. Yih}, \bibinfo{author}{P.~W. Koh}, \bibinfo{author}{M.~Iyyer}, \bibinfo{author}{L.~Zettlemoyer}, \bibinfo{author}{H.~Hajishirzi},
\newblock \bibinfo{title}{{FActScore}: Fine-grained atomic evaluation of factual precision in long form text generation},
\newblock in: \bibinfo{booktitle}{EMNLP}, \bibinfo{year}{2023}. \URLprefix \url{https://arxiv.org/abs/2305.14251}.
\bibitem[{Saad-Falcon et~al.(2024)Saad-Falcon, Khattab, Potts, and Zaharia}]{Falcon2024}
\bibinfo{author}{J.~Saad-Falcon}, \bibinfo{author}{O.~Khattab}, \bibinfo{author}{C.~Potts}, \bibinfo{author}{M.~Zaharia},
\newblock \bibinfo{title}{{ARES}: An automated evaluation framework for retrieval-augmented generation systems},
\newblock in: \bibinfo{booktitle}{Proceedings of the 2024 Conference of the North American Chapter of the Association for Computational Linguistics: Human Language Technologies}, \bibinfo{year}{2024}, pp. \bibinfo{pages}{4392--4408}.
\bibitem[{Qu et~al.(2020)Qu, Yang, Qiu, Croft, Zhang, and Iyyer}]{Qu2020}
\bibinfo{author}{C.~Qu}, \bibinfo{author}{L.~Yang}, \bibinfo{author}{M.~Qiu}, \bibinfo{author}{W.~B. Croft}, \bibinfo{author}{Y.~Zhang}, \bibinfo{author}{M.~Iyyer},
\newblock \bibinfo{title}{{BERT} with history answer embedding for conversational question answering},
\newblock in: \bibinfo{booktitle}{Proceedings of the 43rd International ACM SIGIR Conference}, \bibinfo{year}{2020}, pp. \bibinfo{pages}{1133--1136}.
\bibitem[{Zhang et~al.(2025)Zhang, Liu, Chen, and Wang}]{zhang2025uncertainty}
\bibinfo{author}{W.~Zhang}, \bibinfo{author}{Y.~Liu}, \bibinfo{author}{H.~Chen}, \bibinfo{author}{X.~Wang},
\newblock \bibinfo{title}{To retrieve or not to retrieve? uncertainty detection for dynamic retrieval-augmented generation},
\newblock \bibinfo{journal}{arXiv preprint arXiv:2501.09292}  (\bibinfo{year}{2025}). \URLprefix \url{https://arxiv.org/abs/2501.09292}.

\end{thebibliography}

\appendix

\end{document}